\documentclass{article}
\usepackage[cm]{fullpage}
\usepackage[utf8]{inputenc}
\usepackage{authblk}
\usepackage{epsf}
\usepackage[english]{babel}
\usepackage{xcolor}
\usepackage{amsmath}
\usepackage{amssymb}
\usepackage[super,numbers,sort&compress]{natbib}
\usepackage{graphicx}
\usepackage{lineno}
\usepackage{hyperref}
\usepackage{setspace}
\usepackage{geometry}
\title{Will Artificial Intelligence supersede Earth System and Climate Models?}
\author[1]{Christopher Irrgang}
\author[2,3,4]{Niklas Boers}
\author[5,6,7]{Maike Sonnewald}
\author[8]{Elizabeth A. Barnes}
\author[9]{Christopher Kadow}
\author[10]{Joanna Staneva}
\author[1]{Jan Saynisch-Wagner}
\affil[1]{Helmholtz Centre Potsdam, German Research Centre for Geosciences GFZ, Potsdam, Germany}
\affil[2]{Department of Mathematics and Computer Science, Free University of Berlin, Germany}
\affil[3]{Potsdam Institute for Climate Impact Research, Potsdam, Germany}
\affil[4]{Department of Mathematics and Global Systems Institute, University of Exeter, UK}
\affil[5]{Program in Atmospheric and Oceanic Sciences, Princeton University, Princeton, NJ 08540, USA}
\affil[6]{NOAA/OAR Geophysical Fluid Dynamics Laboratory, Ocean and Cryosphere Division, Princeton, NJ 08540, USA}
\affil[7]{University of Washington, School of Oceanography, Seattle, WA, USA}
\affil[8]{Colorado State University,  Fort Collins, USA}
\affil[9]{German Climate Computing Center DKRZ, Hamburg, Germany}
\affil[10]{Helmholtz-Zentrum Geesthacht, Center for Material and Coastal Research HZG, Geesthacht, Germany}
\date{\today}

\begin{document}
\maketitle

\begin{abstract}
We outline a perspective of an entirely new research branch in Earth and climate sciences, where deep neural networks and Earth system models are dismantled as individual methodological approaches and reassembled as learning, self-validating, and interpretable Earth system model-network hybrids. Following this path, we coin the term "Neural Earth System Modelling'' (NESYM) and highlight the necessity of a transdisciplinary discussion platform, bringing together Earth and climate scientists, big data analysts, and AI experts. We examine the concurrent potential and pitfalls of Neural Earth System Modelling and discuss the open question whether artificial intelligence will not only infuse Earth system modelling, but ultimately render them obsolete.
\end{abstract}

For decades, scientists have utilized mathematical equations to describe geophysical and climate processes and to construct deterministic computer simulations that allow for the analysis of such processes. Until recently, process-based models had been considered irreplaceable tools that helped to understand the complex interactions in the coupled Earth system and that provided the only tools to predict the Earth system's response to anthropogenic climate change. 

The provocative thought that Earth system models (ESMs) might lose their fundamental importance in the advent of novel artificial intelligence (AI) tools has sparked both a gold-rush feeling and contempt in the scientific communities. On the one hand, deep neural networks have been developed that complement and have started to outperform the skill of process-based models in various applications, ranging from numerical weather prediction to climate research. On the other hand, most neural networks are trained for isolated applications and lack true process knowledge. Regardless, the daily increasing data streams from Earth system observation (ESO), increasing computational resources, and the availability and accessibility of powerful AI tools, particularly in machine learning (ML), have led to numerous innovative frontier applications in Earth and climate sciences.
Based on the current state, recent achievements, and recognised limitations of both process-based modelling and AI in Earth and climate research, we draw a perspective on an imminent and profound methodological transformation, hereafter named Neural Earth System Modelling (NESYM). To solve the emerging challenges, we highlight the necessity of new transdisciplinary collaborations between the involved communities.

\section*{Overview on Earth System Modelling and Earth System Observations}
Earth system models (ESMs)\cite{Prinn2013} combine process-based models of the different sub-systems of the Earth system into an integrated numerical model that yields for a given state of the coupled system at time $t$ the tendencies associated with that state, i.e., a prediction of the system state for time $t + 1$. The individual model components, or modules, describe sub-systems including the atmosphere, the oceans, the carbon and other biogeochemical cycles, radiation processes, as well as land surface and vegetation processes and marine ecosystems. These modules are then combined by a dynamic coupler to obtain a consistent state of the full system for each time step.

\begin{figure}
\centering
\includegraphics[width=\linewidth]{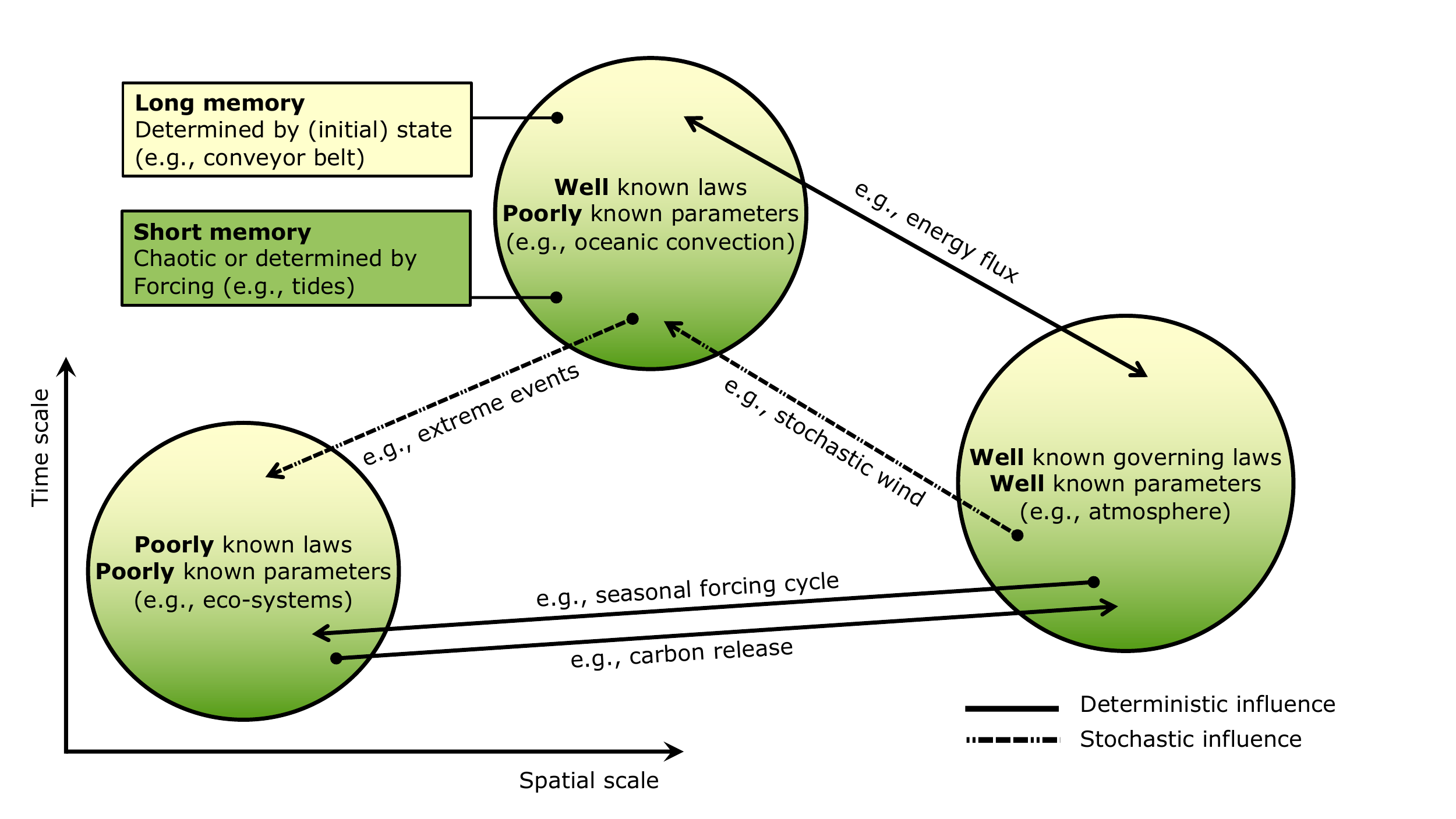}
\caption{Symbolic representation of Earth system components and exemplary deterministic or stochastic coupling mechanisms on long and short spatio-temporal scales.}
\label{fig:esm_sketch}
\end{figure}

For some parts of the Earth system, the primitive physical equations of motion are known explicitly, such as the Navier-Stokes equations that describe the fluid dynamics of the atmosphere and oceans (Fig.~\ref{fig:esm_sketch}). In practise, it is impossible to numerically resolve all relevant scales of the dynamics and approximations have to be made. For example, the fluid dynamical equations for the atmosphere and oceans are integrated on discrete spatial grids, and all processes that operate below the grid resolution have to be parameterised to assure a closed description of the system. Since the multi-scale nature of the dynamics of geophysical fluids implies that the subgrid-scale processes interact with the larger scales that are resolved by the model, (stochastic) parameterization of subgrid-scale processes is a highly non-trivial, yet unavoidable, part of climate modelling \cite{Lin2002,Klein2010,Berner2017}.

For other parts of the Earth system, primitive equations of motion, such as the Navier-Stokes equations for atmospheric motion, do not exist. Essentially, this is due to the complexity of the Earth system, where many phenomena that emerge at a macroscopic level are not easily deducible from microscopic-scales that may or may not be well-understood. A typical example is given by ecosystems and the physiological processes governing the vegetation that covers vast parts of the land surface, as well as their interactions with the atmosphere, the carbon and other geochemical cycles. Also for these cases, approximations in terms of parameterizations of potentially crucial processes have to be made.

Regardless of the specific process, such parameterizations induce free parameters in ESMs, for which suitable values have to be found empirically. The size of state-of-the-art ESMs mostly prohibits systematic calibration methods such as, e.g., the ones based on Bayesian inference, and the models are therefore often tuned manually. The quality of the calibration as well as the overall accuracy of the model can only be assessed with respect to relatively sparse observations of the last 170 years, at most, and there is no way to assess the models’ skill in predicting future climate conditions\cite{Knutti2008}. The inclusion of free parameters possibly causes biases or structural model errors and the example of the discretized spatial grid suggests that the higher the spatial resolution of an ESM, the smaller the potential errors. Likewise, it is expected that the models’ representation of the Earth system will become more accurate the more processes are resolved explicitly. 

The inclusion of a vastly increasing number of processes, together with continuously rising spatial resolution, have indeed led to the development of comprehensive ESMs that have become irreplaceable tools to analyse and predict the state of the Earth system. From the first assessment report of the Intergovernmental Panel on Climate Change (IPCC) in 1990 to the fifth phase of the Climate Model Intercomparison Project (CMIP5)\cite{Taylor2012} and the associated fifth IPCC assessment report in 2014, the spatial resolution has increased from around 500km to up to 70km. In accordance, the CMIP results show that the models have, over the course of two decades, greatly improved in their accuracy to reproduce crucial characteristics of the Earth system, such as the evolution of the global mean temperatures (GMT) since the beginning of instrumental data in the second half of the 19th century, or the average present-day spatial distribution of temperature or precipitation \cite{IPCC,Eyring2016}.

Despite the tremendous success of ESMs, persistent problems and uncertainties remain:

(1) A crucial quantity for the evaluation of ESMs is the equilibrium climate sensitivity (ECS), defined as the amount of equilibrium GMT increase that results from an instantaneous doubling of atmospheric carbon dioxide \cite{Knutti2017}. There remains a large ECS range in current ESM projections and reducing these uncertainties, and hence the uncertainties of future climate projections, is one of the key challenges in the development of ESMs. Nevertheless, from CMIP5 to CMIP6, the likely range of ECS has widened from $2.$1--$4.7^{\circ}$C to $1.8$--$5.6^{\circ}$C\cite{Meehleaba2020,Zelinka2020}. A highly promising line of research in this regard focuses on the identification of emergent constraints, which in principle allow to narrow down the projected range for a model variable of interest, given that the variable has a concise relationship with another model variable that can be validated against past observations \cite{Cox2018,Hall2019}. The development of suitable data-driven techniques for this purpose is still in its infancy.

(2) Both theoretical considerations and paleoclimate data suggest that several sub-systems of the Earth system can abruptly change their state in response to gradual changes in forcing\cite{Lenton2008,Boers2018}. There is concern that current ESMs will not be capable of predicting future abrupt climate changes, because the instrumental era of less than two centuries has not experienced comparable transitions, and model validation against paleoclimate data evidencing such events remains impossible due to the length of the relevant time scales\cite{Valdes2011}. In an extensive search, many relatively abrupt transitions have been identified in future projections of CMIP5 models\cite{Drijfhout2015}, but due to the nature of these rare, high-risk events, the accuracy of ESM in predicting them remains untested.

(3) Current ESMs are not yet suitable for assessing the efficacy or the environmental impact of carbon dioxide removal techniques, which are considered key mitigation options in pathways realizing the Paris Agreement \cite{ipcc2018}. Further, ESMs still insufficiently represent key environmental processes such as the carbon cycle, water and nutrient availability, or interactions between land use and climate. This can impact the usefulness of land-based mitigation options that rely on actions such as biomass energy with carbon capture and storage or nature-based climate solutions \cite{SRCCL2019,SROCC2019}.

(4) The distributions of time series encoding Earth system dynamics typically exhibit heavy tails. Extreme events such as heat waves and droughts, but also extreme precipitation events and associated floods, have always caused tremendous socio-economic damages. With ongoing anthropogenic climate change, such events are projected to become even more severe, and the attribution of extremes poses another outstanding challenge of Earth system science\cite{Otto2015}. While current ESMs are very skilful in predicting average values of climatic quantities, there remains room for improvement in representing extremes.  

In addition to the possible solutions to these fundamental challenges, improvements of the overall accuracy of ESMs can be expected from more extensive and more systematic integration of the process-based numerical models with observational data. Earth system observations (ESOs) are central to ESMs, serving a multitude of purposes. ESOs are used to evaluate and compare process-based model performance, to generate model parameters and initial model states, or as boundary forcing of ESMs \cite{Balsamo2018,Hersbach2020}. ESOs are also used to directly influence the model output by either tuning or nudging parameters that describe unmodeled processes, or by the more sophisticated methods of data assimilation that alter the model's state variables to bring the model output in better agreement with the observations\cite{Evensen2009}. To incorporate uncertainty into  model predictions, variational interfaces have been used\cite{Blei2017}. Existing techniques for assimilating data into ESMs fall into two main categories, each with their own limitations. Gradient-based optimization, as in four-dimensional variational (4DVar) schemes, is the current state of the art for efficiency and accuracy, but currently requires time consuming design and implementation of adjoint calculation routines tailored to each model. Ensemble-based Kalman filter (EnKF) schemes are gradient-free but produce unphysical outputs and rely on strong statistical assumptions that are often unsatisfied, leading to biases and overconfident predictions\cite{Houtekamer2016}. The main problems of contemporary ESM data assimilation are 1) nonlinear dynamics and non-Gaussian error budgets in combination with the high dimensionality of many ESM components \cite{Leeuwen2010,Leeuwen2019,Vetra-Carvalho2018}, and 2) constraining the governing processes over the different spatio-temporal scales found in coupled systems \cite{Penny2019,Browne2019}.  ML approaches can be used to combine the accuracy of 4DVar with the flexibility of EnKF, essentially allowing optimization-based assimilation in cases where gradients are currently unavailable. Furthermore, these traditional approaches of  model and observation fusion have slowly been expanded or replaced by ML methods in recent years \cite{Salcedo-Sanz2020,Irrgang2019,Irrgang2020}. 

ESOs cover a wide range of spatio-temporal scales and types, ranging from a couple of centimeters to tens of thousands of kilometers, and from seconds and decades to millennia. The types of observations range from in-situ measurements of irregular times and spaces (ship cruises, buoy arrays, etc.), over single time series (ice and sediment cores, tide gauges, etc.) to satellite-based global 2D or 3D data fields (altimetry, gravimetry, radio occultation, etc.). 
The amount of available observations is rapidly increasing and has reached a threshold where automated analysis becomes crucial. Yet, the available observational data pool still contains large gaps in time and space that prevent building a holistic observation-driven picture of the coupled Earth system, which result from insufficient spatio-temporal data resolution, too short observation time periods, and largely unobserved compartments of Earth systems like, for instance, abyssal oceans. The combination of these complex characteristics render Earth system observations both challenging and particularly interesting for AI applications.

\section*{From Machine Learning-based Data Exploration Towards Learning Physics}
ML and other AI techniques have achieved stunning results in computer vision \cite{Voulodimos2018}, speech and language models \cite{Brown2020}, medical science \cite{Loh2018,Topol2019}, economical and societal analytics \cite{Nosratabadi2020}, and other disciplines \cite{Hagerty2019,Perc2019}. Due to this wide-spread integration into both fundamental research and end-user products, and despite shortcomings and inherent limitations \cite{Papernot2016,Adadi2018,Walton2018,Geirhos2020}, ML is already praised as a key disruptive technology of the 21st century \cite{Girasa2020}.
In contrast, the usage of ML in Earth and climate sciences is still in its infancy. A key observation is that ML concepts from computer vision and automated image analysis can be isomorphically transferred to ESO imagery. Pioneering studies demonstrated the feasibility of ML for remote sensing data analysis, classification tasks, and parameter inversion already in the 1990s \cite{Dawson1992,Miller1995,Serpico1996,Hsieh1998a}, and climate-model emulation in the early 2000s \cite{Knutti2003}.
The figurative Cambrian explosion of AI techniques in Earth and climate sciences, however, only began over the last five years and will rapidly continue throughout the coming decades.

Under the overarching topic of ESO data exploration, ML has been applied for a huge variety of statistical and visual use cases. Classical prominent examples are pattern recognition in geo-spatial observations, climate data clustering, automated remote-sensing data analysis, and time series prediction \cite{Lary2016,Salcedo-Sanz2020}. In this context, ML has been applied across various spatial and temporal scales, ranging from short-term regional weather prediction to Earth-spanning climate phenomena. Significant progress has been made in developing purely data-driven weather prediction networks, which start to compete with process-based model forecasts \cite{Arcomano2020,Weyn2019,Weyn2020}. ML contributed to the pressing need to improve the predictability of natural hazards, for instance, by uncovering global extreme-rainfall teleconnections \cite{Boers2019}, or by improving long-term forecasts of the El Ni\~no Southern Oscillation (ENSO)\cite{Ham2019,Yan2020}.
ML-based image filling techniques were utilized to reconstruct missing climate information, allowing to correct previous global temperature records \cite{Kadow2020}. Furthermore, ML was applied to analyze climate data sets, e.g., to extract specific forced signals from natural climate variability \cite{Barnes2019,Barnes2020} or to predict clustered weather patterns \cite{Chattopadhyay2020}. In these applications, the ML tools function as highly specialized agents that help to uncover and categorize patterns in an automated way. A key methodological advantage of ML in comparison to covariance-based spatial analysis lies in the possibility to map nonlinear processes  \cite{Ramachandran2017,Lu2018}. At the same time, such trained neural networks lack actual physical process knowledge, as they solely function through identifying and generalizing statistical relations by minimizing pre-defined loss measures for a specific task \cite{Rumelhart1986}. Consequently, research on ML in Earth and climate science differs fundamentally from the previously described efforts of advancing ESMs in terms of methodological development and applicability.

Concepts of utilizing ML not only for physics-blind data analyses, but also as surrogates and methodological extensions for ESMs have only recently started to shape \cite{Reichstein2019}. Scientists started pursuing the aim that ML methods learn aspects of Earth and climate physics, or at least plausibly relate cause and effect. The combination of ML with process-based modelling is the essential distinction from the previous ESO data exploration. Lifting ML from purely diagnosis-driven usage towards the prediction of geophysical processes will also be crucial for aiding climate change research and the development of mitigation strategies \cite{Huntingford2019}.

Following this reasoning, ML methods can be trained with process-based model data to inherit a specific geophysical causation or even emulate and accelerate entire forward simulations. For instance, ML has been used in combination with ESMs and ESOs to invert space-borne oceanic magnetic field observations to determine the global ocean heat content \cite{Irrgang2019}. Similarly, a neural network has been trained with a continental hydrology model to recover high-resolution terrestrial water storage from satellite gravimetry \cite{Irrgang2020}. ML plays an important role for upscaling unevenly distributed carbon flux measurements to improve global carbon monitoring systems\cite{Jung2020}. As such, the eddy covariance technique was combined with ML to measure the net ecosystem exchange of $\text{CO}_2$ between ecosystems and the atmosphere, offering a unique opportunity to study ecosystem responses to climate change \cite{Tramontana2020}. ML has shown remarkable success in representing subgrid-scale processes and other parameterizations of ESMs, given that sufficient training data were available. As such, neural networks were applied to approximate turbulent processes in ocean models \cite{Bolton2019} and atmospheric subgrid processes in climate models \cite{Rasp2018}. Several studies highlight the potential for ML-based parameterization schemes \cite{OGorman2018,Gagne2020,Han2020,Beucler2020,Yuval2020}, 
helping step-by-step to gradually remove numerically and human-induced simplifications and other biases of ESMs \cite{Brenowitz2018}.

While some well-trained ML tools and simple hybrids have shown higher predictive power than traditional state-of-the-art process-based models, only the surface of new possibilities, but also of new scientific challenges, has been scratched. So far, ML, ESMs, and ESO have largely been independent tools. Yet, we have reached the understanding that applications of physics-aware ML and model-network hybrids pose huge benefits by filling up niches where purely process-based models persistently lack reliability.

\section*{The Fusion of Process-based Models and Artificial Intelligence}
\begin{figure}
\centering
\includegraphics[width=\linewidth]{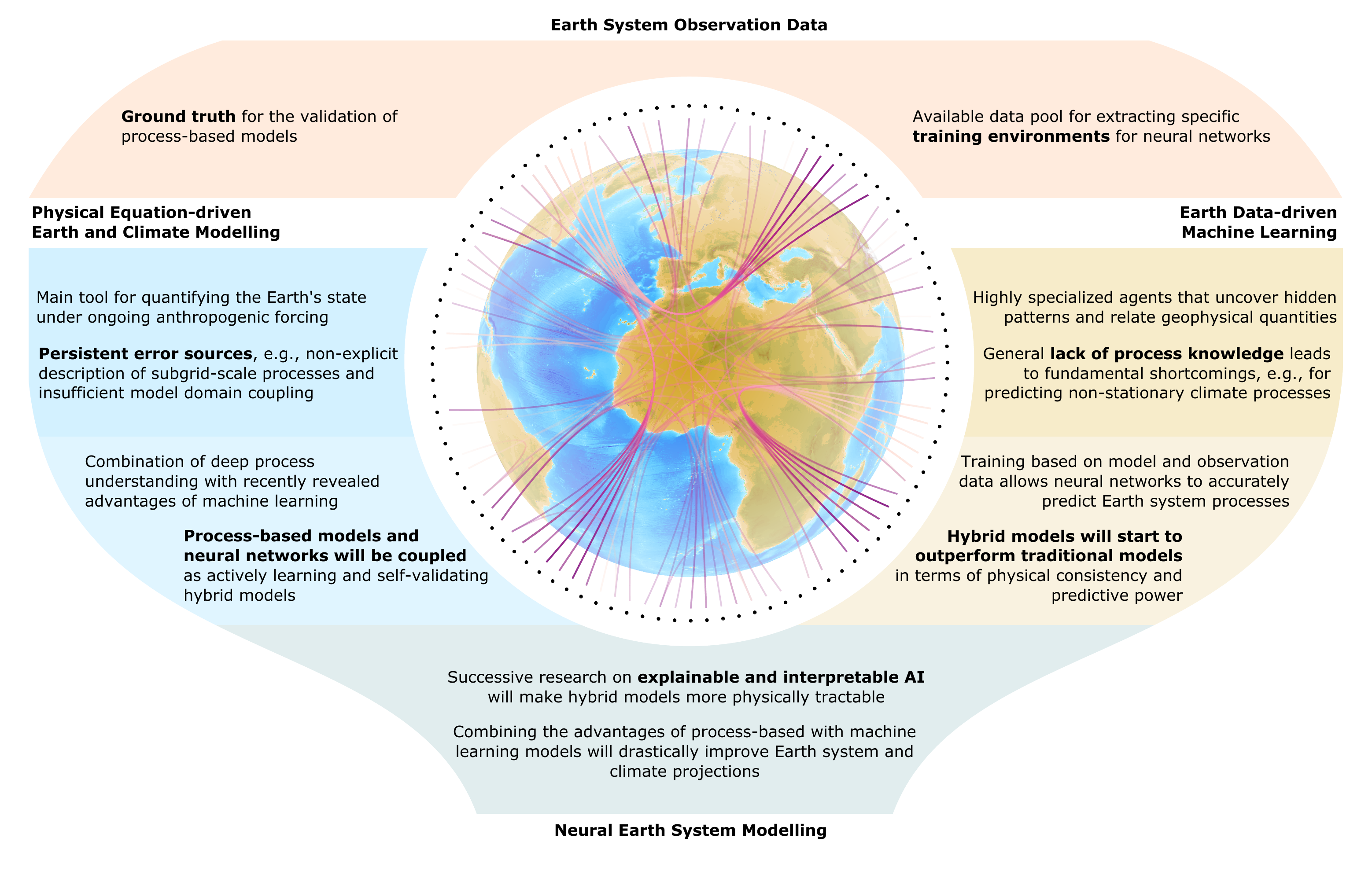}
\caption{Successive stages of the fusion process of Earth system models and artificial intelligence.}
\label{fig:fusion_sketch}
\end{figure}
The idea of hybrids of process-based and ML models is not new \cite{Krasnopolsky2006}. So far, hybrids have almost exclusively been thought of as numerical models that are enhanced by ML to either improve the models’ performance in the sense of a useful metric, or to accelerate the forward simulation time in exchange for a decrease in simulation accuracy. Along with the general advance regarding the individual capabilities and limitations of ESMs and ML methods, respectively, also the understanding of how ML can enhance process-based modelling has evolved. This progress allows ML to take over more and more components of ESMs, gradually blending the so far strict distinction between process-based modelling and data-driven ML approaches. Even more so, entirely new methodological concepts are dawning that justify acknowledging Neural Earth System Modelling as a distinct research branch (Fig.~\ref{fig:fusion_sketch}).

The long-term goal will be to consistently integrate the recently discovered advantages of ML into the already decade-long source of process knowledge in Earth system science. However, this evolution does not come without methodological caveats, which need to be investigated carefully. For the sake of comparability, we distinguish between weakly coupled NESYM hybrids, i.e., an ESM or AI technique benefits from information from the respective other, and strongly coupled NESYM hybrids, i.e., fully coupled model-network combinations that dynamically exchange information between each other.

The emergent development of weak hybrids is predominantly driven by the aim to resolve the previously described ESM limitations, particularly unresolved and especially sub-grid scale processes. Neural networks can emulate such processes after careful training with simulation data from a high-resolution model that resolves the processes of interest, or with  relevant ESO data. The next methodological milestone will be the integration of such trained neural networks into ESMs for operational usage. First tests have indicated that the choice of the AI technique, e.g., neural networks versus random forests, seems to be crucial for the implementation of learning parameterization schemes, as they can significantly deteriorate the ESM’s numerical stability \cite{Brenowitz2020}. Thus, it is not only important to identify how neural networks can be trained to resolve ESM limitations, but also how such ML-based schemes can be stabilized in the model physics context and how their effect on the process-based simulation can be evaluated and interpreted \cite{Brenowitz2020a}. The limitations of ML-based parameterization approaches can vary widely for different problems or utilized models and, consequently, should be considered for each learning task individually \cite{Seifert2020}. Nevertheless, several ideas have been proposed to stabilize ML parameterizations, e.g., by enforcing physical consistency through customized loss functions in neural networks and specific network architectures \cite{Beucler2019,Beucler2020}, or by optimizing the considered high-resolution model training data \cite{Yuval2020}. In addition, an ESM blueprint has been proposed, in which learning parameterizations can be targeted through searching an optimal fit of statistical measures between ESMs, observations, and high-resolution simulations \cite{Schneider2017}. In this context, further efforts have been made to enhance an ESM not with ML directly, but in combination with a data assimilation system \cite{Evensen2009}. For instance, emulating a Kalman filter scheme with ML has been investigated \cite{Cintra2014, Wahle2015}, an ML-based estimation of atmospheric forcing uncertainties used as error covariance information in data assimilation has been proposed \cite{Irrgang2020b}, as well as further types of Kalman-network hybrids \cite{Brajard2020,Ruckstuhl2020}.

In the second class of weak hybrids, the model and AI tasks are transposed, such that the information flow is directed from the model towards the AI tool. Here, neural networks are trained directly with model state variables, their trajectories, or with more abstract information like seasonal signals, interannual cycles, or coupling mechanisms. The goal of the ML application might not only be model emulation, but also inverting non-linear geophysical processes \cite{Irrgang2019}, learning geophysical causation \cite{Runge2019}, or predicting extreme events \cite{Boers2014,Qi2020}.
In addition to these inference and generalization tasks, a key question in this sub-discipline is whether a neural network can learn to outperform the utilized process-based trainer model in terms of physical consistency or predictive power. ESOs play a vital role in this context, as they can serve as additional training constraints for a neural network training, allowing it to build independent self-evaluation measures \cite{Irrgang2020}.

The given examples generally work well for validation and prediction scenarios within the given training distribution. Out-of-distribution samples, in contrast, pose a huge challenge for supervised learning, which renders the “learning from the past'' principle particularly ill-posed for prediction tasks in NESYM. As a consequence, purely data-driven AI methods will not be able to perform accurate climate projections on their own, because of the both naturally and anthropogenically induced non-stationarity of the climate and Earth system. Overcoming these limitations requires a deeper holistic integration in terms of strongly coupled hybrids and the consideration of further, less constrained training techniques like unsupervised training \cite{Sonnewald2020} and generative AI methods \cite{Leinonen2019,Gagne2020,Stengel2020}. For example, problems of pure AI methods with non-stationary training data can be attenuated by combining them with physical equations describing the changing energy-balance of the Earth system due to anthropogenic greenhouse-gas emissions \cite{Huber2012}.
In addition, first steps towards physics-informed AI have been made by ML-based and data-driven discovery of physical equations \cite{Zanna2020a} and by the implementation of neural partial differential equations \cite{Lagaris1998,Raissi2019} into the context of climate modelling \cite{Ramadhan2020}.

Continuous maturing of the methodological fusion process will allow building hybrids of neural networks, ESMs and ESOs that dynamically exchange information. ESMs will soon utilize output from supervised and unsupervised neural networks to optimize their physical consistency and, in turn, feed back improved information content to the ML component. ESOs form another core element and function as constraining ground truth of the AI-infused process prediction. Similar to the adversarial game of generative networks \cite{Goodfellow2014}, or coupling mechanisms in an ESM \cite{Hurrell2013}, also strongly coupled NESYM hybrids will require innovative interfaces that control the exchange of information that are, so far, not available. In addition, we formulate key characteristics and goals of this next stage: 

(1) Hybrids can better reproduce and predict out-of-distribution samples and extreme events,

(2) hybrids perform constrained and consistent simulations that obey physical conservation laws despite potential shortcomings of the hybrids’ individual components,

(3) hybrids include integrated adaptive measures for self-validation and self-correction, and

(4) NESYM allows replicability and interpretability.

We believe that cross-discipline collaborations between Earth system and AI scientists will become more important than ever to achieve these milestones. Frontier applications of Neural Earth system models are manifold. Yet, ultimately, NESYM hybrids need to drastically improve the current forecast limits of geophysical processes and contribute towards understanding the Earth's susceptible state in a changing climate. Consequently, not only the fusion of ESM and AI will be in the research focus, but also AI interpretability and resolving the common notion of a black box (Fig.~\ref{fig:metagraph}).

\section*{Peering into the Black Box}
\begin{figure}
\centering
\includegraphics[width=0.9\linewidth]{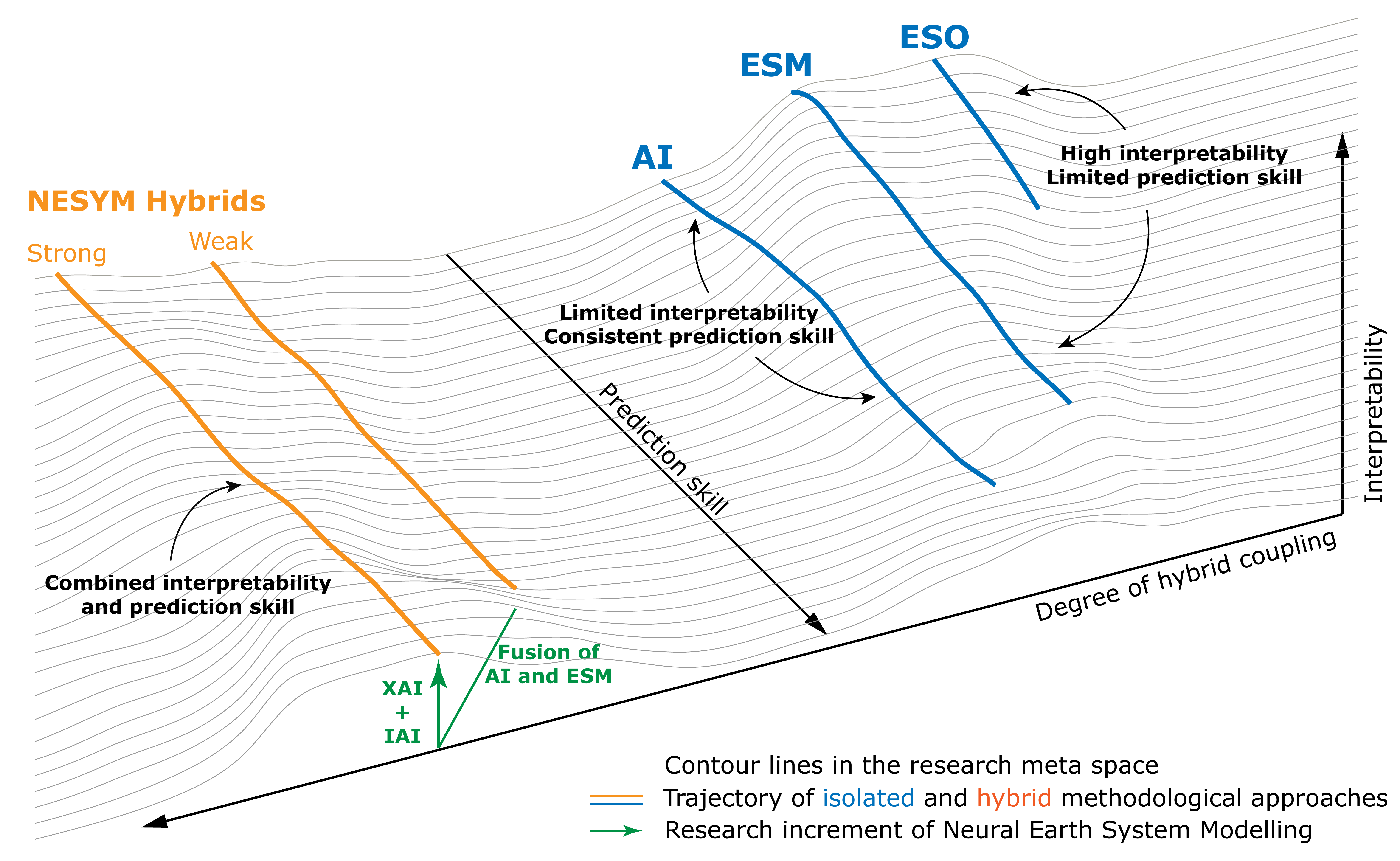}
\caption{Qualitative comparison of isolated (AI - Artificial Intelligence, ESM - Earth System Model, ESO - Earth System Observation) and hybrid methodological approaches. The respective approaches are represented as trajectories in a meta space of hybrid coupling degree, interpretability, and prediction skill. The goal of Neural Earth System Modelling (NESYM) is to integrate the interpretability and to exceed the prediction skill of the respective isolated approaches. In this meta space, the necessary research increment to achieve this goal can be described through an increase in the degree of hybrid coupling (Fusion of AI and ESM) and an increase in interpretability (XAI - explainable AI, IAI - interpretable AI). }
\label{fig:metagraph}
\end{figure}

ML has emerged as a set of methods based on the combination of statistics, applied mathematics and computer science, but it comes with a unique set of hurdles. Peering into the black box and understanding the decision making process of the ML method, termed explainable AI (XAI), is critical to the use of these tools. Especially in the physical sciences, adaptation of ML suffers from a lack of interpretability, particularly supervised ML. In contrast and in addition to XAI stands the call for interpretable AI (IAI), i.e., building specifically interpretable ML models from the beginning on, instead of explaining ML predictions through post-process diagnostics \cite{Rudin2019}.

Ensuring that what is ‘learned’ by the machine is physically tractable or causal, and not due to trivial coincidences \cite{Balaji2020}, is important before ML tools are used, e.g., in an ESM setting targeted at decision making. Thus, explainability provides the user with trust in the ML output, improving its transparency. This is critical for ML use in the policy-relevant area of climate science as society is making it increasingly clear that understanding the source of AI predictive skill is of crucial importance\cite{AIethicsEU,AIethicsUS}. Ensuring the ML method is getting the right answers for the right reasons is essential given the transient nature of the climate system. As the climate continues to respond to anthropogenic climate change, NESYM will be required to make predictions of continually evolving underlying distributions and XAI/IAI will be critical to ensuring that the skill of the ML method can be explained, and inspire trust in its extrapolation to future climate regimes. There are many ML tools at our disposal, and XAI can assist researchers in choosing the optimal ML architecture, inputs, outputs, etc. By analyzing the decision making process, climate scientists will be able to better incorporate their own physical knowledge into the ML method, ultimately leading to improved results. Perhaps least appreciated in geoscientific applications thus far is the use of XAI to discover new science \cite{Toms2020}. When the ML method is capable of making a prediction, XAI allows us to ask “what did it learn?”. In this way, ML becomes more than just a prediction and allows scientists to ask “why?” as they normally would, but now with the power of ML.

Explaining the source of an ML applications skill can be done retrospectively\cite{Rudin2019}. The power of XAI for climate and weather applications has very recently been demonstrated \cite{McGovern2019,Toms2020,Ebert-Uphoff2020}. For example, neural networks coupled with the XAI attribution method known as layerwise relevance propagation (LRP)\cite{Olden2004,Bach2015} have revealed modes of variability within the climate system, sources of predictability across a range of timescales, and indicator patterns of climate change \cite{Toms2020,Barnes2020}. There is also evidence that XAI methods can be used to evaluate climate models against observations, identifying the most important climate model biases for the specific prediction task \cite{Barnes2020b}. However, these methods are in their infancy and there is vast room for advancements in their application, making it explicitly appropriate to employ them within the physical sciences.

Unsupervised ML can be intuitively IAI through the design of experiments. For example, applying clustering on closed model budgets of momentum ensures all relevant physics are represented, and can be interpreted in terms of the statistically dominant balances. In this manner, different regimes can be discovered \cite{Sonnewald2019,Sonnewald2020}. Adversarial learning has been an effective tool for generating super-resolution fields of atmospheric variables in climate models \cite{Stengel2020}. Furthermore, unsupervised ML approaches have been proposed for discovering and quantifying causal interdependencies and dynamical links inside a system, such as the Earth’s climate \cite{Runge2019,Runge2019b}. The development of ESMs is increasingly turning to process-oriented diagnostics (PODs)\cite{Maloney2019}, where a certain process is targeted and used as a benchmark for model improvement. A revolution of analysis tools has been called for, and ML is poised to be part of this change \cite{Eyring2019,Schlund2020,Reichstein2019}. For instance, the POD approach has been applied to evaluate the ability of ESM projections to simulate atmospheric interactions and to constrain climate projection uncertainties \cite{Nowack2020}.

Given the importance of both explainability and interpretability for improving ML generalization and scientific discovery, we need climate scientists working together with AI scientists to develop methods that are tailored to the field’s needs. This is not just an interesting exercise - it is essential for the proper use of AI for NESYM development and use (Fig.~\ref{fig:metagraph}). Earth and climate scientists can aid the development of consistent benchmarks that allow evaluating both stand-alone ML and NESYM hybrids in terms of geophysical consistency \cite{Rasp2020}. However, help of the AI community is needed to resolve other recently highlighted ML pitfalls, for instance, translating the concepts of adversarial examples and deep learning artifacts \cite{Buckner2020} into the ESM context or finding new measures to identify and avoid shortcut learning \cite{Geirhos2020} in NESYM hybrids. In summary, only combined efforts and continuous development of both ESM and AI will evolve Neural Earth System Modelling.

Our perspective should not only be seen as the outline of a  promising scientific pathway to achieve a better understanding of the Earth’s present and future state, but also as an answer to the recent support call from the AI community \cite{Rolnick2019}. Based on the recent advances in applying AI to Earth system and climate sciences, it seems to be a logical succession that AI will take over more and more tasks of traditional statistical and numerical ESM methods. Yet, in its current stage, it also seems unthinkable that AI alone can solve the climate prediction problem. In the forthcoming years, AI will necessarily need to rely on the geophysical determinism of process-based modelling and on careful human evaluation. However, once we find solutions to the foreseeable limitations described above and can build interpretable and geophysically consistent AI tools, this next evolutionary step will seem much more likely.

\section*{Acknowledgements}
This study was funded by the Helmholtz Association and by the Initiative and Networking Fund of the Helmholtz Association through the project Advanced Earth System Modelling Capacity (ESM). NB acknowledges funding by the Volskwagen foundation and the European Union’s Horizon 2020 research and innovation program under grant agreement No 820970 (TiPES).

\section*{Authors' contributions}
CI conceived the paper and organized the collaboration. All authors contributed to writing and revising all chapters of this manuscript. In particular, NB and CI drafted the ESM overview, JSW and JS drafted the ESO and DA overview, CI and CK drafted the chapter 'From Machine Learning-based Data Exploration Towards Learning Physics', CI and JSW and NB drafted the chapter 'The Fusion of Process-based Models and Artificial Intelligence', MS and EB and CI drafted the chapter 'Peering into the Black Box'.

\section*{Competing Interest}
The authors declare no competing interest.

\bibliographystyle{naturemag}

\end{document}